\documentclass{llncs}
\usepackage{makeidx}  
\usepackage{amsmath,amsfonts,amssymb}
\usepackage{graphicx}
\usepackage[tight]{subfigure}
\usepackage{url}
\usepackage{float}
\usepackage{hyperref}
\usepackage{comment}
\usepackage{cite}

\usepackage{color}

\begin{document}
\frontmatter        
\pagestyle{headings}  
\mainmatter             

\title{Projection Inpainting Using Partial Convolution for Metal Artifact Reduction}
\titlerunning{Deep Learning For Limited Angle Tomography}  
%
\author{Lin~Yuan\inst{1}\and
Yixing~Huang\inst{1} \and 
Andreas~Maier\inst{1,2}
}

\institute{ Friedrich-Alexander Universit\"at Erlangen-N\"urnberg, 91058 Erlangen, Germany\\
\and
Erlangen Graduate School in Advanced Optical Technologies
(SAOT), 91058 Erlangen, Germany}

\authorrunning{Lin Yuan et al.} 
%
%
\institute{Pattern Recognition Lab, Friedrich-Alexander Universit\"at Erlangen-N\"urnberg, 91058 Erlangen, Germany\\
\and
Erlangen Graduate School in Advanced Optical Technologies
(SAOT), 91058 Erlangen, Germany}

\maketitle              

\begin{abstract}


In computer tomography, due to the presence of metal implants in the patient body, reconstructed images will suffer from metal artifacts. In order to reduce metal artifacts, metals are typically removed in projection images. Therefore, the metal corrupted projection areas need to be inpainted. For deep learning inpainting methods, convolutional neural networks (CNNs) are widely used, for example, the U-Net. However, such CNNs use convolutional filter responses on both valid and corrupted pixel values, resulting in unsatisfactory image quality. In this work, partial convolution is applied for projection inpainting, which only relies on valid pixels values. The U-Net with partial convolution and conventional convolution are compared for metal artifact reduction. Our experiments demonstrate that the U-Net with partial convolution is able to inpaint the metal corrupted areas better than that with conventional convolution.



\keywords{Deep learning, partial convolution, metal artefact reduction}
\end{abstract}
\renewcommand{\baselinestretch}{0.9}

\section{Introduction}


In computed tomography (CT), image patients may contain metal implants such as a dental filling, spine screws, surgical clips or artificial hips. In this situation, reconstructed images will suffer from metal artifacts, typically in the form of strong bright and dark streaks. These artifacts are caused by various effects, including most prominently beam hardening, noise, scattering and the non-linear partial volume effect. Metals have much higher attenuation values than body tissues, leading to severe beaming hardening effect \cite{brooks1976beam}. X-ray photon flux follows a Poisson distribution. Due to high absorption of photons by metals, a low photon-count X-ray beam causes relatively high Poisson noise and detector electronic noise. In addition, metal implants usually have well defined boundaries, causing the nonlinear partial volume effect \cite{glover1980nonlinear}.

Various metal artifact reduction (MAR) algorithms have been proposed \cite{gjesteby2016metal}. Among them, projection inpainting methods are the most common methods, such as linear interpolation \cite{kalender1987reduction}, polynomial interpolation \cite{wei2004x}, wavelet domain interpolation \cite{zhao2000x} and sinusoidal curve fitting \cite{liu2005metal}. Reprojection methods \cite{crawford1988high,naidu2004method,bal2006metal,prell2009novel,karimi2012segmentation} are also widely used for the inpainting of metal corrupted projections, where the metal region in a initial reconstruction is replaced by a tissue-class model, where soft tissue is most frequently used. In addition, data normalization is beneficial for projection interpolation. With this idea, the normalized MAR (NMAR) algorithms are proposed \cite{meyer2010normalized,meyer2011adaptive}.

Recently, powerful deep learning methods using convolutional neural networks (CNNs) have also been applied for projection inpainting in MAR applications \cite{gjesteby2017deep,zhang2018convolutional,ghani2018deep,ghani2019fast,gottschalk2019deep}. However, such CNNs use convolutional filter responses on both valid and corrupted pixel values, resulting in unsatisfactory image quality. In \cite{pcon5}, a partial convolution method has been proposed for image inpainting in the field of computer vision, where the convolution operations only rely on valid pixels, given valid pixel masks. With sufficient successive updates, valid pixel region grows while the invalid blank regions gradually get inpainted. In this work, we introduce this partial convolution method for projection inpainting in the application of MAR. Note that working independently, a multi-domain MAR method using partial convolution has already been proposed in \cite{pimkin2019multi}.

\section{Materials and Methods}
\subsection{Partial convolution}

The concept of partial convolution is actually a general CNN with masks.  A partial convolution layer consists of a partial convolution operation and a mask update function. The partial convolution operation can be expressed as \cite{pcon5},

\begin{equation}
x'=\begin{cases}\boldsymbol{W}^{T}\left( \boldsymbol{X}\odot \boldsymbol{M}\right) \dfrac {\text{sum}(\boldsymbol{1})}{\text{sum}\left( \boldsymbol{M}\right) }+b,\text{ if } \text{sum}\left( \boldsymbol{M}\right) >0,\\ 0,\text{     otherwise   }\end{cases},
\label{Eqn:PCDefinition}
\end{equation}

where $\boldsymbol{W}$ represents the weight of a convolutional layer filter,
$b$ is bias,
$\boldsymbol{X}$ is the pixel or feature of the current convolution (sliding) window,
$\boldsymbol{M}$ is a corresponding binary mask composed of 0 and 1. After each partial convolution operation is completed, the mask undergoes a round of updates. This means that if the convolution can adjust its output on at least one valid input, the mask is switched to 0 at that location. 
$\odot$ indicates element-wise multiplication. 
sum($\boldsymbol{1}$)/sum($\boldsymbol{M}$) is a scaling factor where $\boldsymbol{1}$ has the same shape as $\boldsymbol{M}$, that is, applying an appropriate scaling ratio to adjust the amount of change in the unmasked input. From the definition, we can see that the output of partial convolution only depends on the unmasked input values.

After each partial convolution, the mask is updated \cite{pcon5},
\begin{equation}
m'=\begin{cases}1,\text{ if } \text{sum}\left( M\right) >0,\\ 0,\text{     otherwise,   }\end{cases}
\label{Eqn:maskGrow}
\end{equation}
where $m'$ is the value of the mask at the convolution output pixel. That is to say, if the convolution was able to condition its output on at least one valid input value, then we mark that location to be valid \cite{pcon5}. With the increase of the number of network layers, the pixel value of the mask output 0 is getting less and less, the area of the valid area in the output result is getting larger and larger, and the influence of the mask on the overall loss will also become smaller. The result of this is that when the network is deep enough, all the pixels in the mask will become 1 \cite{pcon5}.

\begin{figure}[H]
\centering
\begin{minipage}[b]{0.38\linewidth}
\centering
\includegraphics[width = \linewidth]{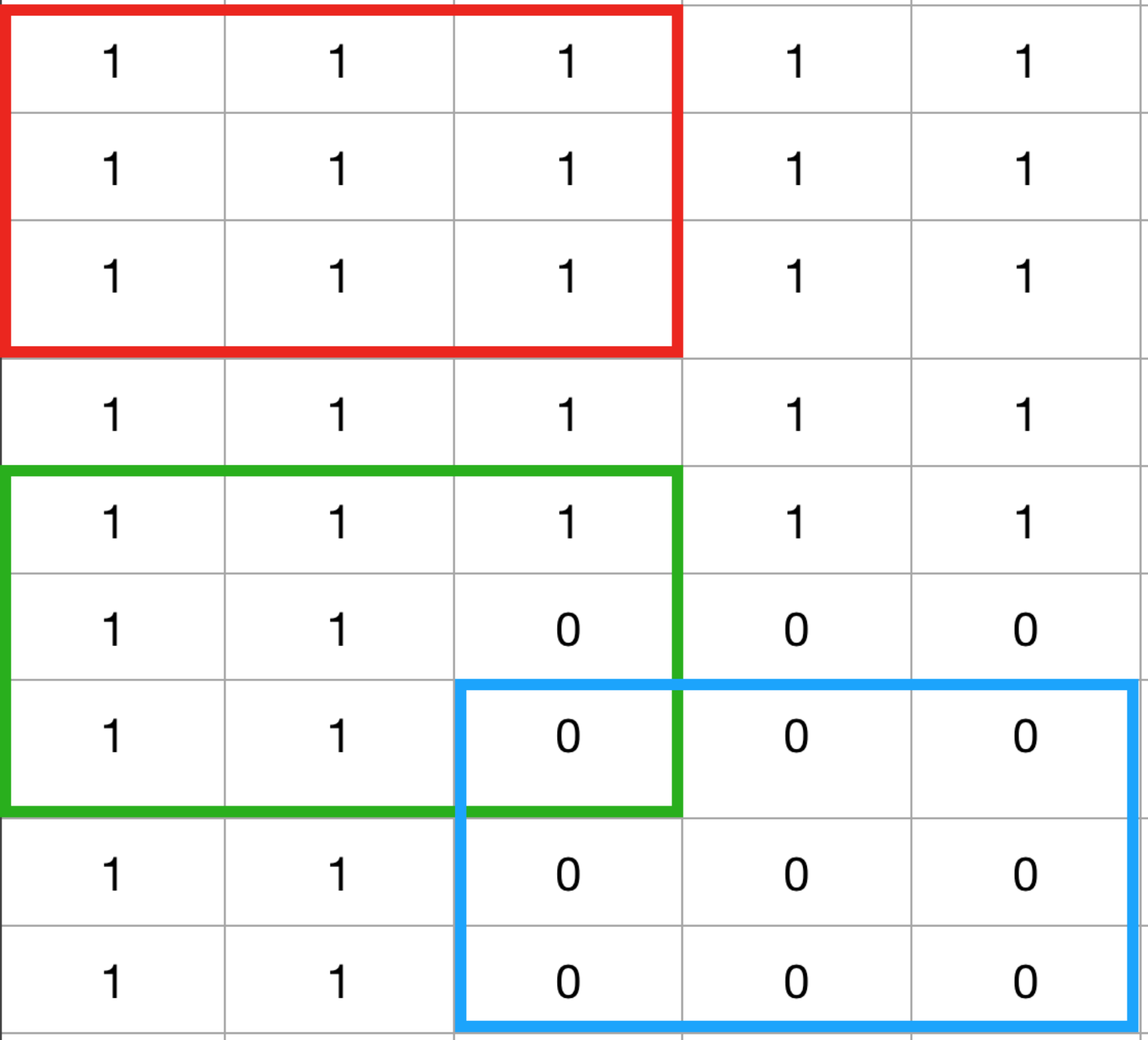}
\caption{mask example
}
\label{Fig:form}
\end{minipage}
\end{figure}

To illustrate the partial convolution, an example is displayed in Fig.~\ref{Fig:form}.

\textbf{Red frame:} At this time, the mask values in the kernel are all 1 (all are correct pixels and need not be filled). Then it will execute the formula of ``if sum ($\boldsymbol{M}$)$>$ 0" in Eqn.~(\ref{Eqn:PCDefinition}). Since all pixels are valid here, the convolution can be processed normally using the conventional convolution.

\textbf{Green frame:}
Although the mask values of the kernel are 0 at the bottom right corner (representing a hole), we can learn something through the nearby normal pixels with 1s.

\textbf{Blue frame:}
At this time, the mask values in the kernel are all 0 (ball pixels need to be filled).
We will not deal with it at the beginning, until there is more information to be passed to the later layer.

In a partial convolution neural network, for regions like the above green frame, the mask $\boldsymbol{M}$ will gradually fill from 0 to 1 because of the formula ``if sum ($\boldsymbol{M}$)$>$ 0" in Eqn.~(\ref{Eqn:maskGrow}). Through this way to the end, a mask of all 1s is obtained, i.e., the entire image has been inpainted, although the intensity values need to be further improved.
\subsection{U-Net architecture} 

In this work, the U-Net \cite{Ronneberger2015U} with conventional convolutions and partial convolutions are compared for projection inpainting in the application of MAR. The general U-Net architecture consists of two symmetrical parts: the previous part of the network is the same as the ordinary convolutional network, using 3x3 convolution and pooled down-sampling, which can grasp the relationship between pixels in the image; the latter part of the network is basically symmetrical with the previous one, but with skip connections for multi-scale feature extraction.

\begin{figure}[H]
\centering
\begin{minipage}[b]{1\linewidth}
\centering
\includegraphics[width = 0.62\linewidth]{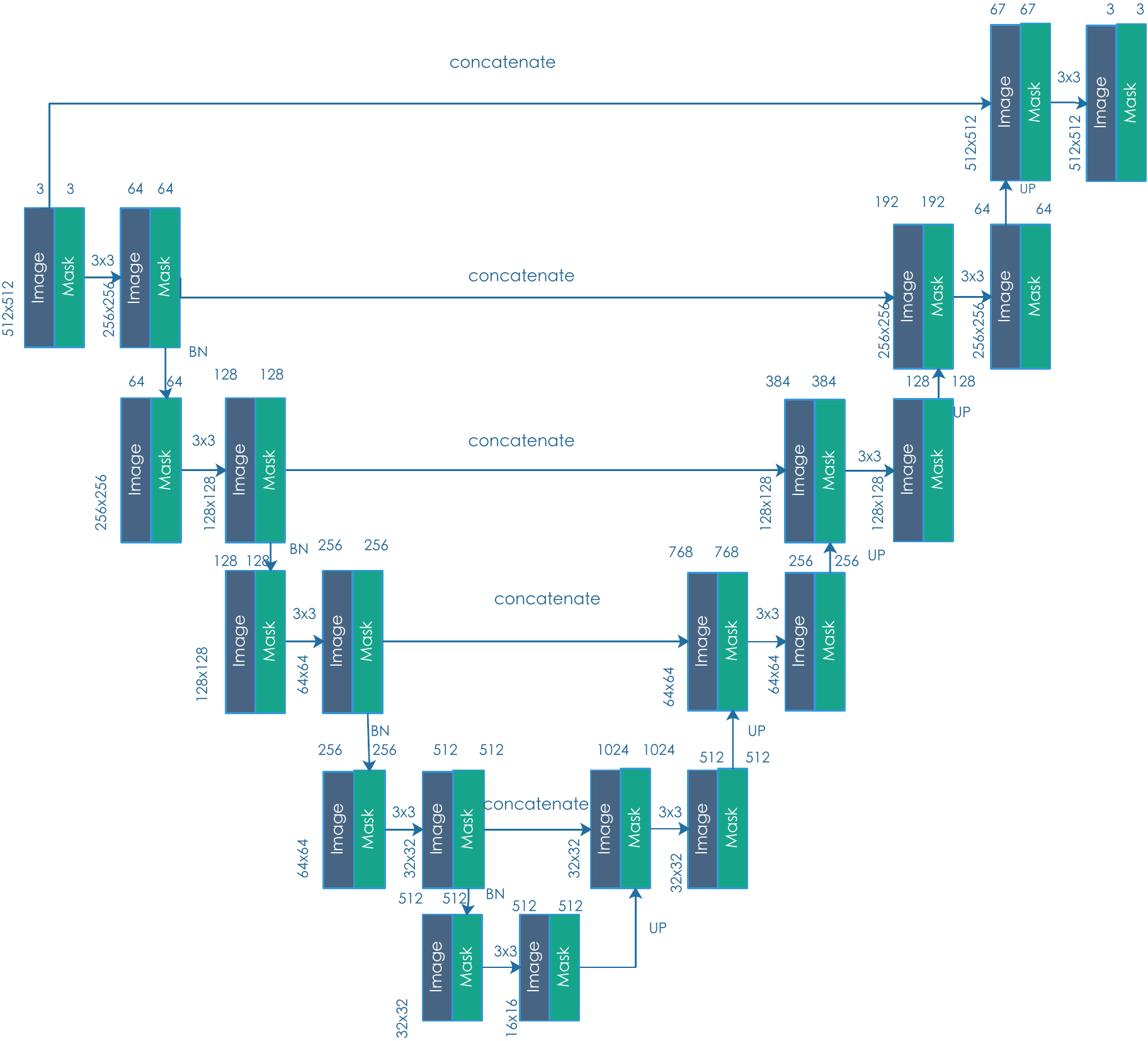}
\end{minipage}
\caption{The architecture of the U-Net with partial convolution.}
\label{Fig:pcon architecture}
\end{figure}


The architecture of the U-Net with partial convolution is displayed in Fig.~\ref{Fig:pcon architecture}, where masks are introduced to the original U-Net architecture. The left part includes 5 partial convolutional layers PCONV1-PCONV5, responsible for image encoding. And the right side includes 5 partial convolutional layers PCONV6-PCONV10 responsible for decoding. 

\subsection{Experimental Setup}
\begin{figure}[htb]
\centering
\begin{minipage}[b]{1\linewidth}
\centering
\includegraphics[width = \linewidth]{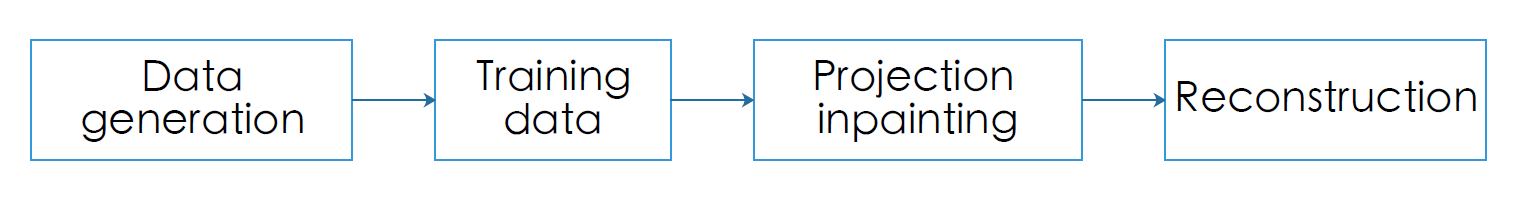}
\caption{Training Processing}
\label{Fig:process}
\end{minipage}
\end{figure}

The flowchart of our experimental setup is shown in Fig.~\ref{Fig:process}, which consists of the following steps.

\begin{figure}[htb]
\centering
\begin{minipage}[b]{0.5\linewidth}
\centering
\includegraphics[width = \linewidth]{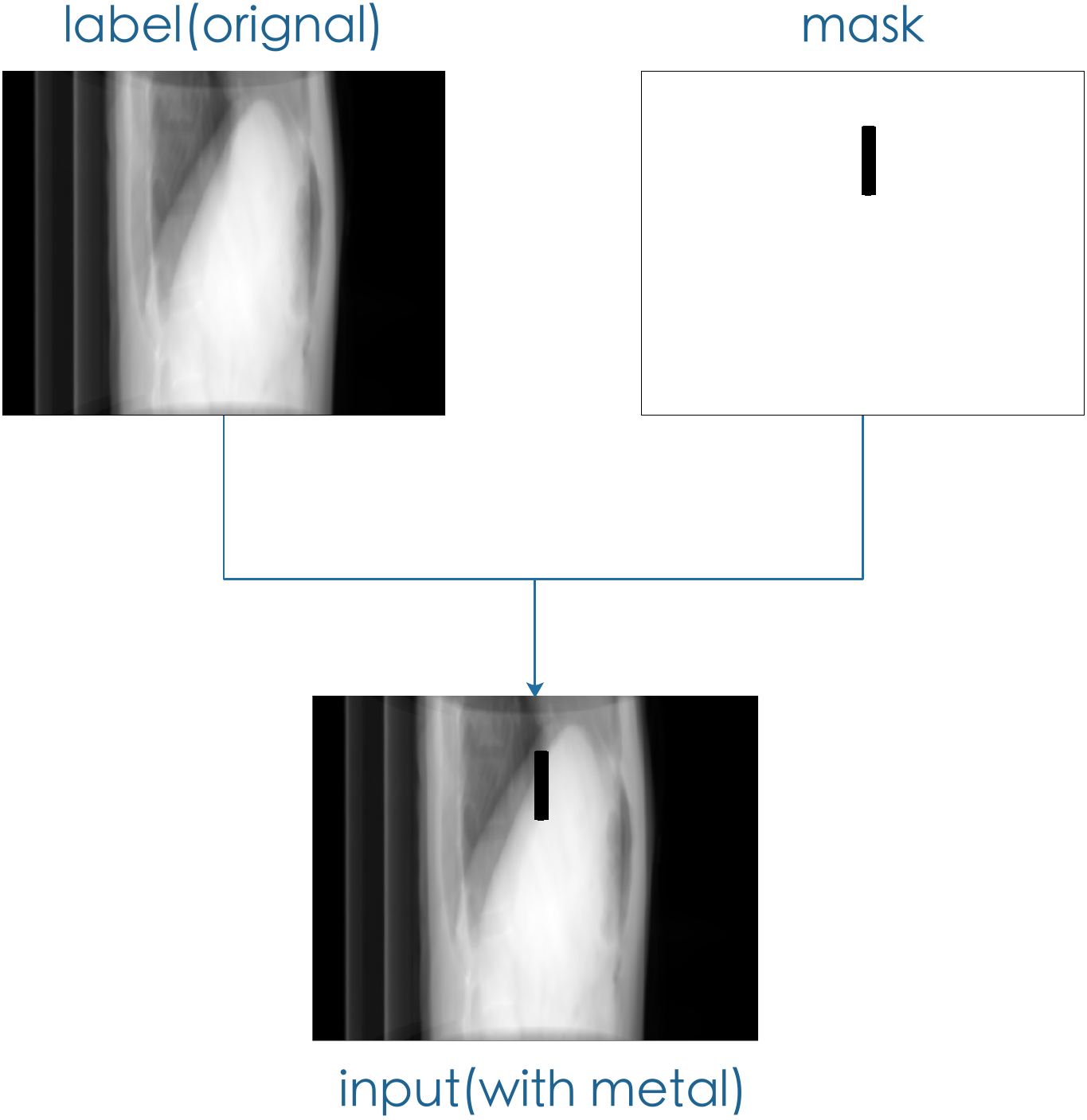}
\end{minipage}
\caption{The generation of metal corrupted projections using an original metal-free projection and a mask.}
\label{Fig:data generation}
\end{figure}

\textbf{Data generation:} We generate original metal-free projections by forward projection of 18 patients' CT volumes on CONRAD  \cite{CONRAD} using an angular step of $1^\circ$. For metal corrupted projections, we randomly generate metals of different sizes and positions in the patient CT volumes and forward project the volumes with metals to get metal corrupted projections. The masks are obtained by forward projection of the metals only. With the above procedures, we have the original projection images, the masks and the projections with metals, as shown in Fig.~\ref{Fig:data generation}. 

\begin{figure}[htb]
\centering
\begin{minipage}[b]{0.8\linewidth}
\centering
\includegraphics[width = \linewidth]{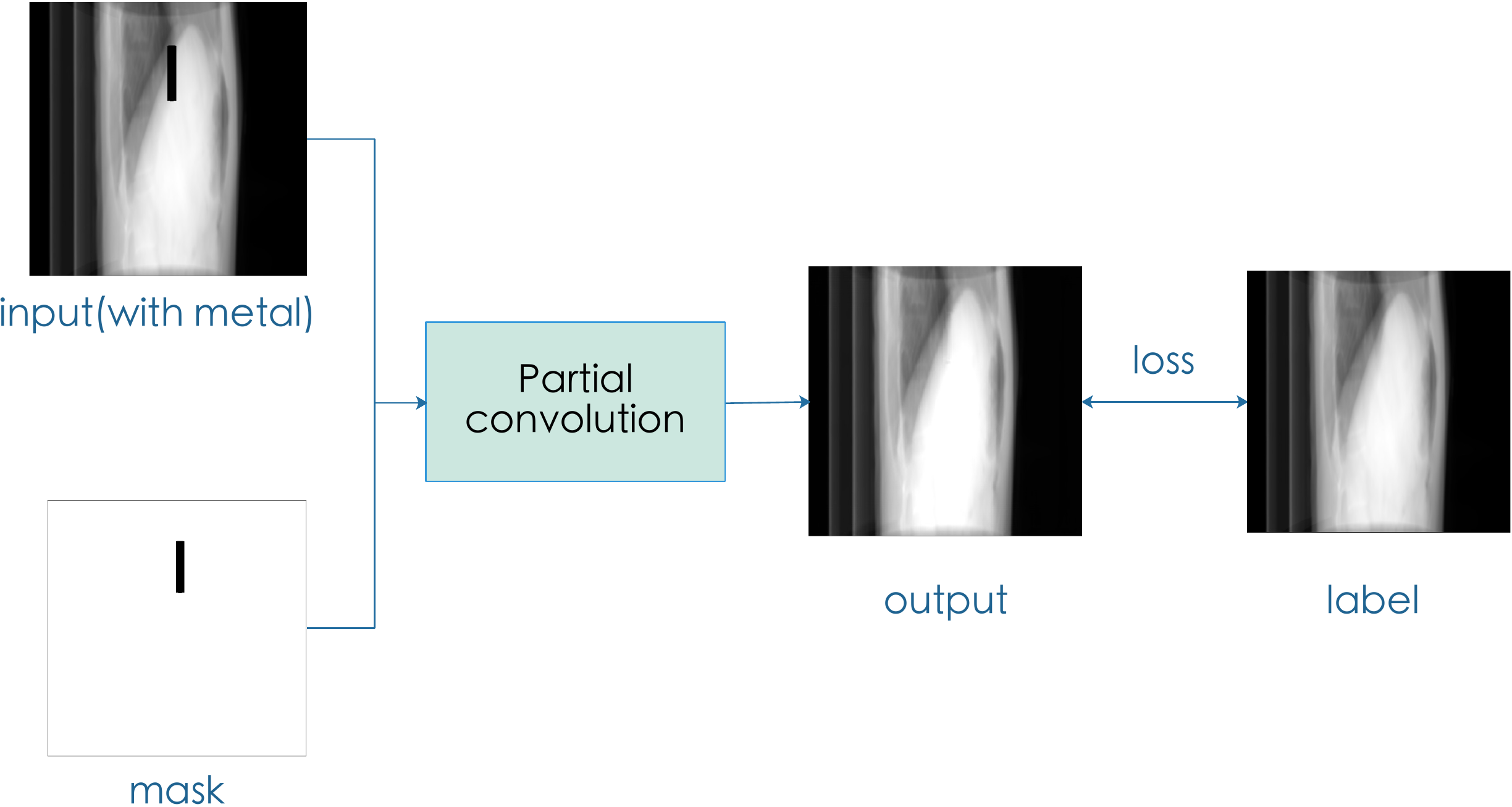}
\caption{Training Process of partial convolution }
\label{Fig:partial con}
\end{minipage}
\end{figure}

\begin{figure}[htb]
\centering
\begin{minipage}[b]{0.8\linewidth}
\centering
\includegraphics[width = \linewidth]{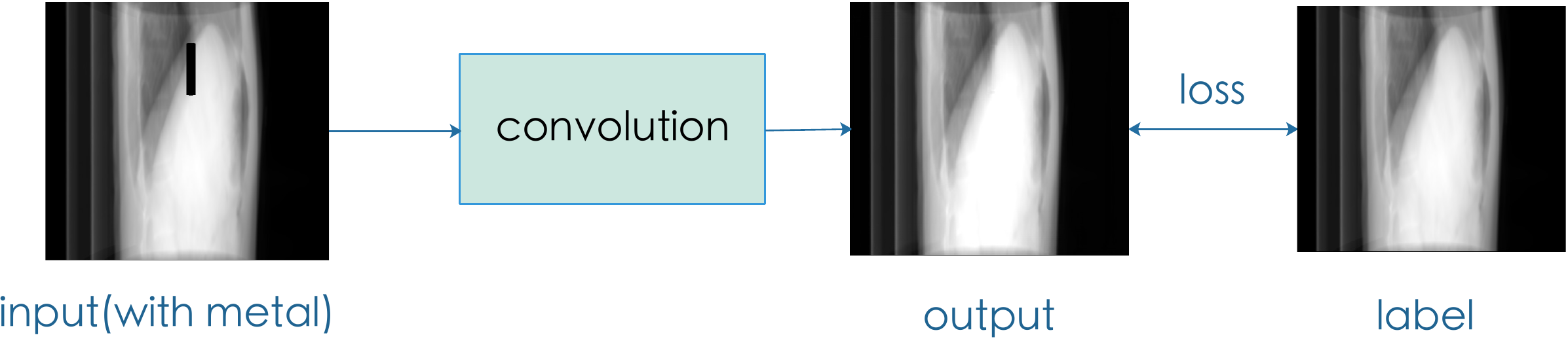}
\end{minipage}
\caption{Training Process of normal convolution}
\label{Fig:unet}
\end{figure}

\textbf{Training data:} We used the pytorch-inpainting-with-partial-conv code on github \cite{python}. On this basis, we have made several modifications for our application. The picture is properly scaled to 512 * 512. Because the original code runs slowly, we changed the batch size to 2. The training process is shown in Fig.~\ref{Fig:partial con}. The input is a metal corrupted projection with its corresponding mask. After partial convolution, the output is obtained and compared with the original image for correction. We use the same loss function as that in \cite{pcon5}. The U-Net with conventional convolution is also applied to inpaint metal corrupted projections. Its process is shown in Fig.~\ref{Fig:unet}.

\textbf{Projection inpainting:} Use the trained model to generate metal-free projections for all the $360^\circ$ projections.

\textbf{Image reconstruction:} CT reconstruction from inpainted projections using the FDK algorithm.

\section{Results And Discussion}
\begin{figure}[htb]
\centering
\begin{minipage}[t]{0.28\linewidth}
\subfigure[]{
\includegraphics[width = \linewidth]{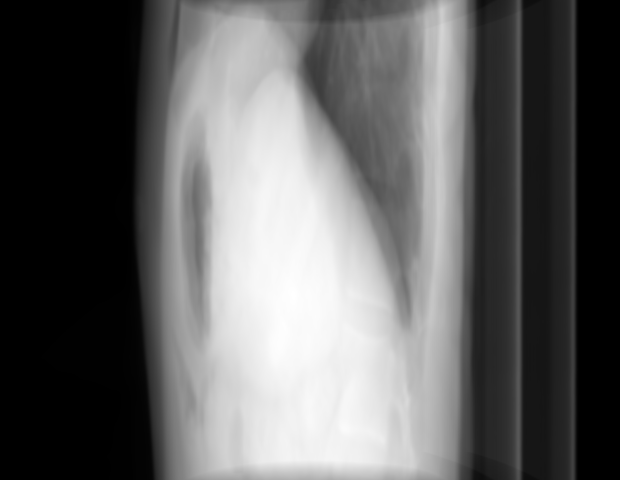}
}
\end{minipage}
\begin{minipage}[t]{0.288\linewidth}
\subfigure[]{
\includegraphics[width = \linewidth]{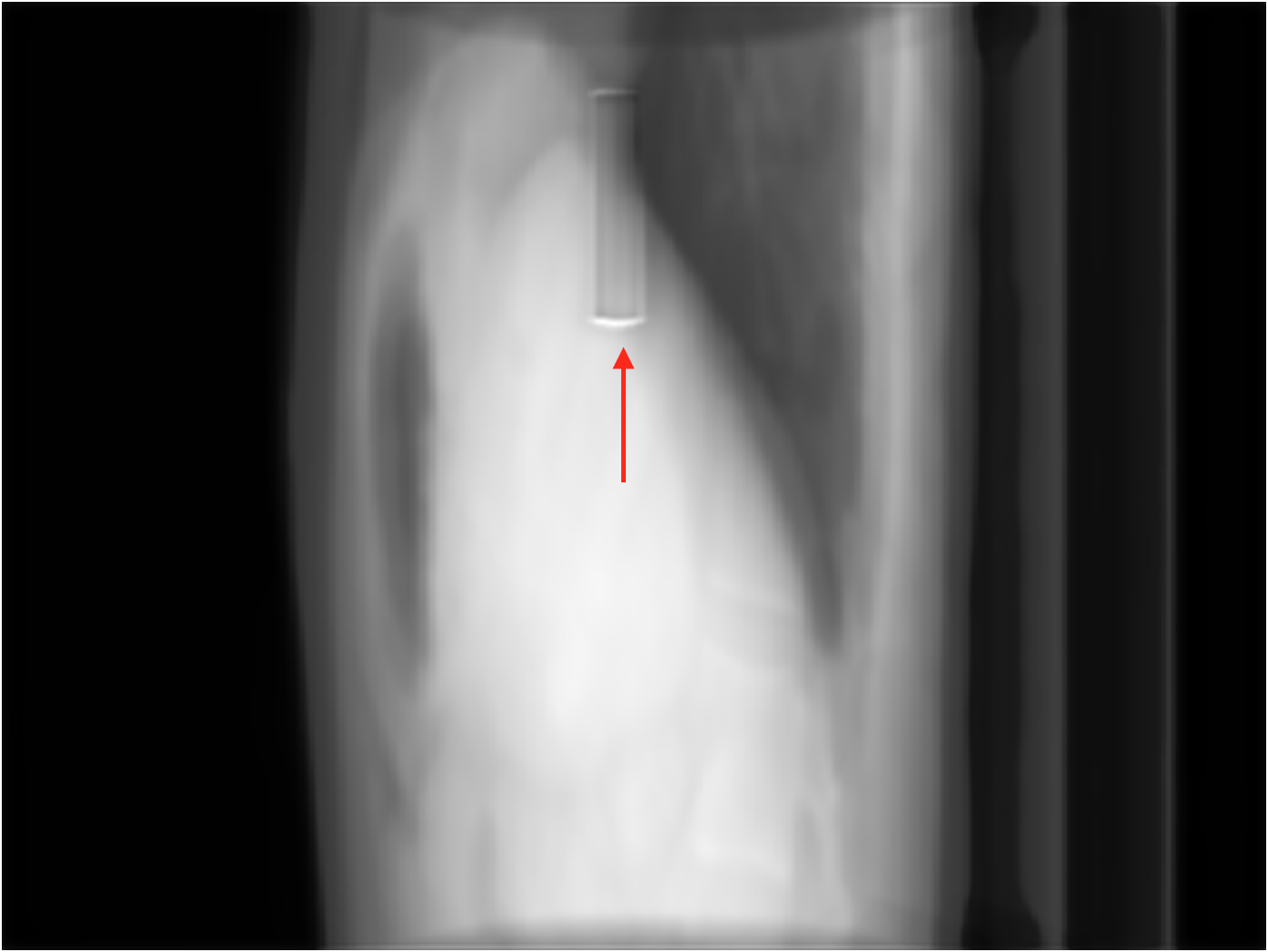}
}
\end{minipage}
\begin{minipage}[t]{0.28\linewidth}
\subfigure[]{
\includegraphics[width = \linewidth]{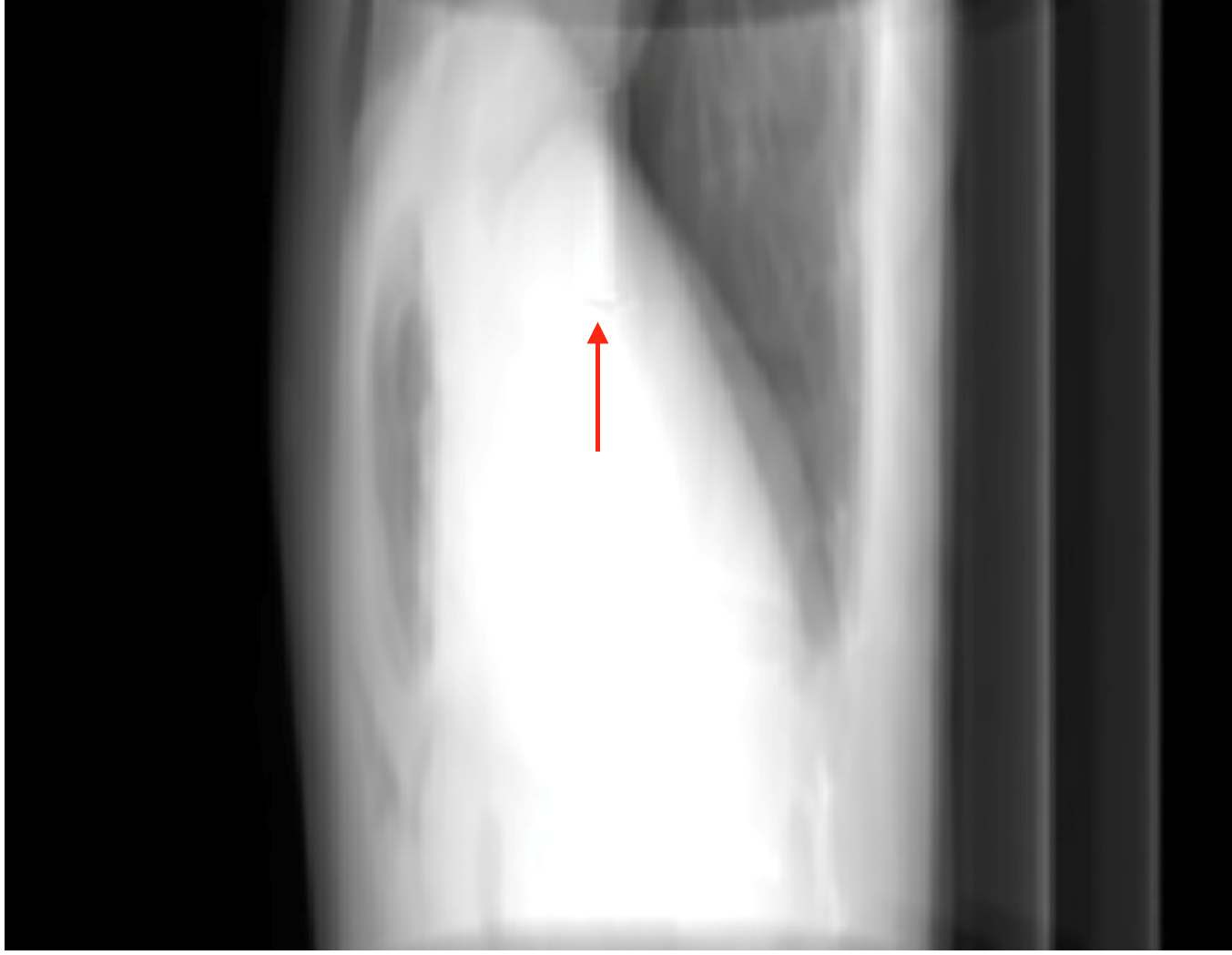}
}
\end{minipage}

\caption{The results of inpainted projection (90th projection): (a) original image; (b) inpainted projection by the U-Net with conventional convolution; (c) inpainted projection by the U-Net with partial convolution.}
\label{Fig:result1}
\end{figure}

The inpainting results of one projection are displayed in Fig.~\ref{Fig:result1}. Fig.~\ref{Fig:result1}(a) is the original image. The red arrows in Figs.~\ref{Fig:result1}(b) and (c) indicate the inpainted regions. Fig.~\ref{Fig:result1}(b) shows that the U-Net with conventional convolution is able to inpaint something in the metal corrupted region. However, er can observe obvious boundaries. Fig.~\ref{Fig:result1}(c) demonstrates that the region inpainted by the U-Net with partial convolution has superior quality, since no obvious boundaries are observed. This demonstrates the benefit of partial convolution for projection inpainting.

\begin{figure}[htb]
\centering
\begin{minipage}[t]{0.22\linewidth}
\subfigure[]{
\includegraphics[width = \linewidth]{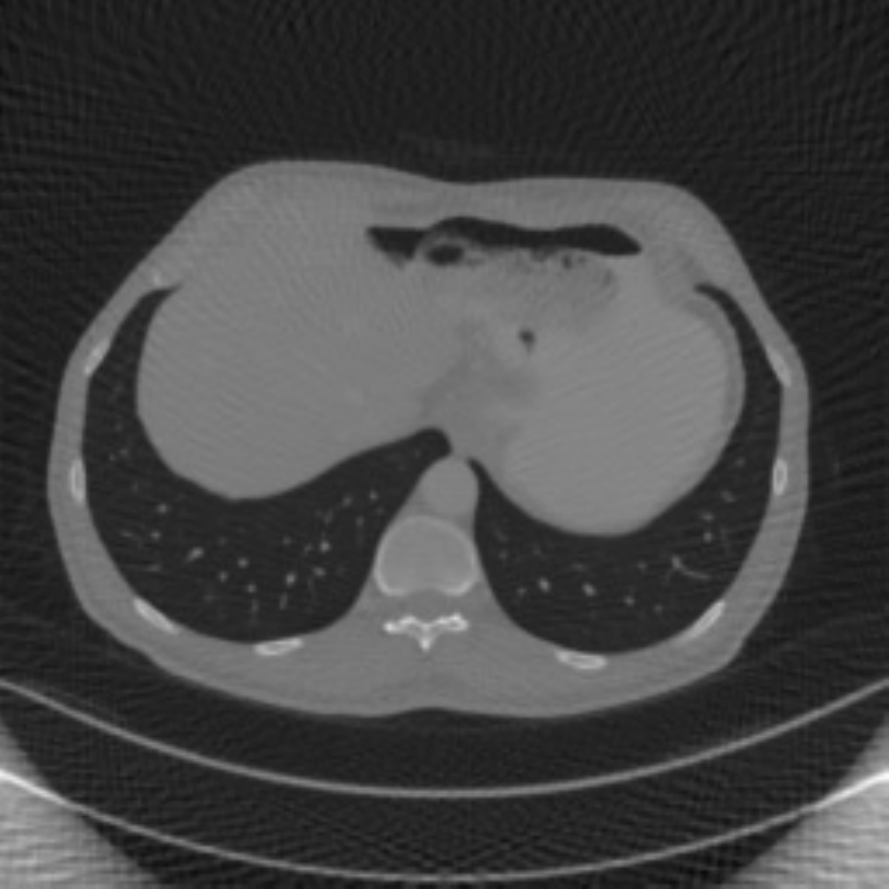}
}
\end{minipage}
\begin{minipage}[t]{0.22\linewidth}
\subfigure[]{
\includegraphics[width = \linewidth]{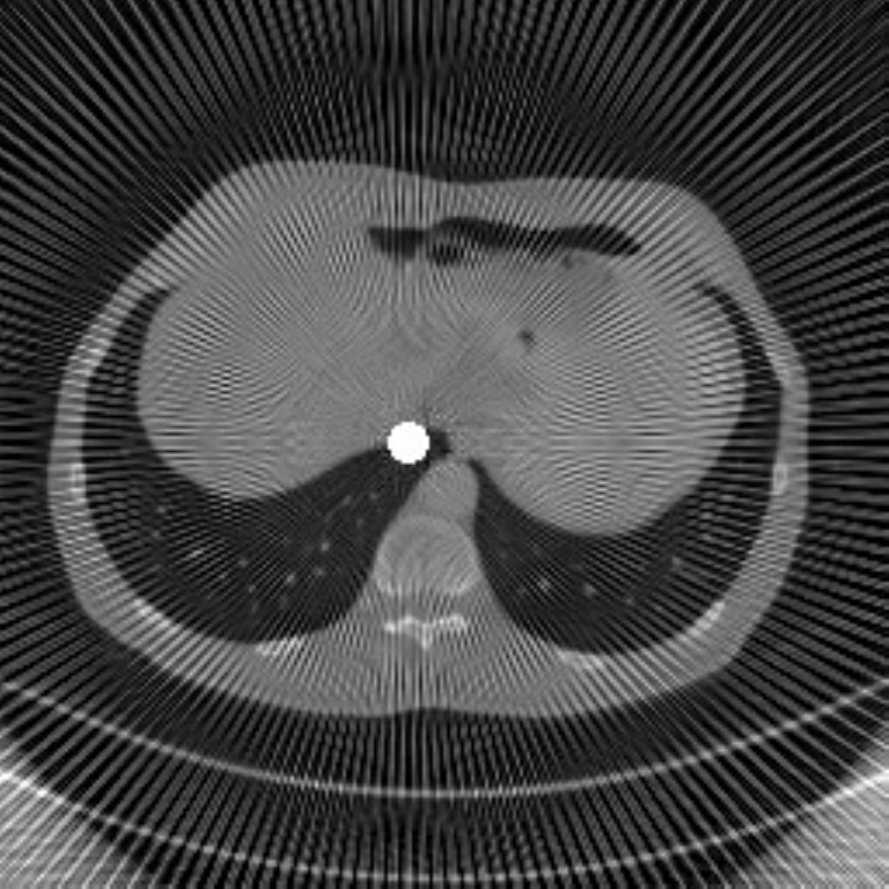}
}
\end{minipage}
\begin{minipage}[t]{0.22\linewidth}
\subfigure[]{
\includegraphics[width = \linewidth]{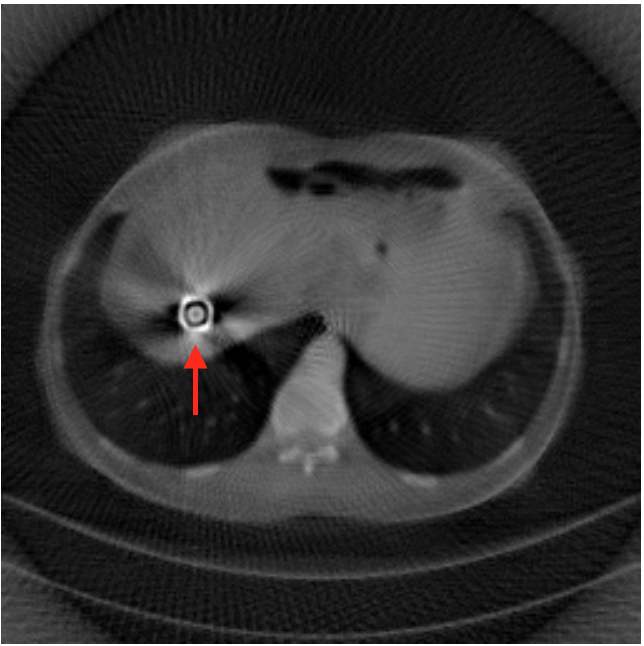}
}
\end{minipage}
\begin{minipage}[t]{0.22\linewidth}
\subfigure[]{
\includegraphics[width = \linewidth]{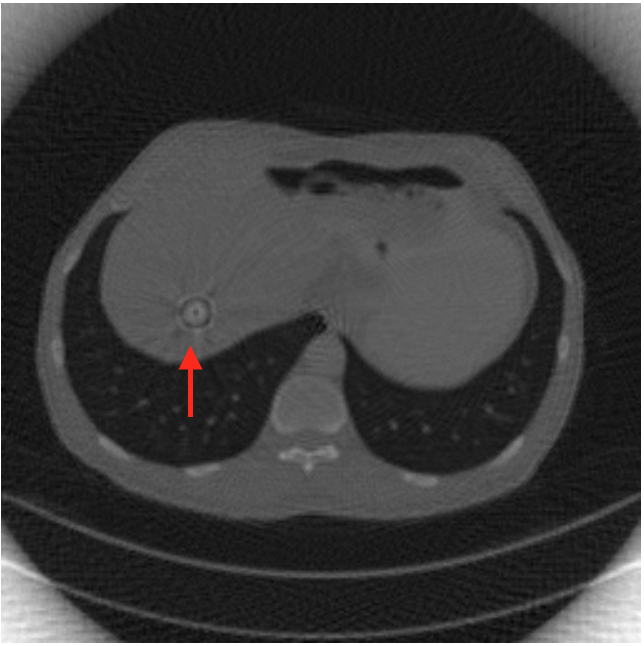}
}
\end{minipage}
\caption{The results of one example slice reconstructed from inpainted projections: (a) ground truth slice; (b) reconstruction directly from metal corrupted projections (with metal in the projections); (c) reconstruction from inpainted projections by the U-Net with conventional convolution; (d) reconstruction from inpainted projections by the U-Net with partial convolution. }
\label{Fig:result2}
\end{figure}

The reconstruction results from inpainted projections are displayed in Fig.~\ref{Fig:result2}. Fig.~\ref{Fig:result2}(a) is the ground truth image. Fig.~\ref{Fig:result2}(b) shows the direct reconstruction from metal corrupted projections. The metal is present in the image with severe radial streak artifacts. Fig.~\ref{Fig:result2}(c) is the reconstruction from inpainted projections by the U-Net with conventional convolution, where most streak artifacts are reduced very well. However, at the position of the metal, the boundary of the metal is presented with artifacts. This is because in the projections the boundary areas are not well inpainted, as displayed in Fig.~\ref{Fig:result1}(b). Fig.~\ref{Fig:result2}(d) shows the reconstruction from inpainted projection by the U-Net with partial convolution. The radial steak artifacts are very well reduced. Moreover, at the position of the metal, the remaining artifacts are much fewer than those in (c). This demonstrates the advantage of partial convolution over conventional convolution in projection inpainting in the application of MAR.

\section{Conclusion}

In this paper, we investigate the application of partial convolution in projection inpainting for MAR. Compared with the U-Net with conventional convlution, the U-Net with partial convolution is able to inpaint the metal corrupted areas better, especially at the boundary areas. As a result, in the reconstruction from inpainted projections by the U-Net with partial convolution, radial streak artifacts are well reduced and the structures near the metal position are well observed.

%
%

%

\end{document}